\documentclass[10pt,twocolumn,letterpaper]{article}

\usepackage[pagenumbers]{cvpr} 

\usepackage[dvipsnames]{xcolor}

\usepackage{times}
\usepackage{epsfig}
\usepackage{graphicx}
\usepackage{amsmath}
\usepackage{amssymb}
\usepackage{multirow}
\usepackage{lipsum}
\usepackage{adjustbox}
\usepackage{booktabs}
\usepackage{makecell}
\usepackage{tabu}
\usepackage{bm}
\usepackage{xcolor}
\usepackage{comment}
\usepackage{colortbl}
\usepackage{color}
\usepackage{pifont}
\usepackage{colortbl}
\definecolor{Gray}{gray}{0.95}
\definecolor{Grayy}{gray}{0.6}

\definecolor{cvprblue}{rgb}{0.21,0.49,0.74}
\usepackage[pagebackref,breaklinks,colorlinks,citecolor=cvprblue]{hyperref}

\def\paperID{4003} 
\def\confName{CVPR}
\def\confYear{2024}

\usepackage{pifont}
\title{Hierarchical Memory for Long Video QA}

\author{
    Yiqin Wang\textsuperscript{1}$^,$\footnotemark[1]~,
    Haoji Zhang\textsuperscript{1}$^,$\footnotemark[1]~,
    Yansong Tang\textsuperscript{1}$^,$\footnotemark[2]~,
    Yong Liu\textsuperscript{1},
    Jifeng Dai\textsuperscript{2},
    Jiashi Feng\textsuperscript{3},
    Xiaojie Jin\textsuperscript{3}$^,$\footnotemark[2]~\\    
    \textsuperscript{1} Shenzhen International Graduate School, Tsinghua University\\
    \textsuperscript{2} Department of Electronic Engineering, Tsinghua University\ \ \ 
    \textsuperscript{3} ByteDance Inc.\\
    {\tt \small \{yq-wang23@mails.,zhj24@mails.,tang.yansong@sz.\}tsinghua.edu.cn}\\
    {\tt \small jinxiaojie@bytedance.com}\\
}

\begin{document}


\maketitle

\def\thefootnote{$*$}\footnotetext{Equal contribution.}
\def\thefootnote{$\dagger$}\footnotetext{Corresponding author.}

\begin{abstract}
    This paper describes our \textbf{champion solution} to the \href{https://sites.google.com/view/loveucvpr24/track1}{LOVEU Challenge @ CVPR'24, Track 1 (Long Video VQA)}.
    Processing long sequences of visual tokens is computationally
    expensive and memory-intensive, making long video question-answering a challenging task.
    The key is to compress visual tokens effectively, reducing memory footprint and decoding latency, while
    preserving the essential information for accurate question-answering.
    We adopt a hierarchical memory mechanism named STAR Memory, proposed in \href{https://invinciblewyq.github.io/vstream-page/}{Flash-VStream}~\cite{flashvstream},
    that is capable of processing long videos with limited GPU memory (VRAM).
    We further utilize the video and audio data of MovieChat-1K training set to fine-tune
    the pretrained weight released by Flash-VStream, achieving 1st place in the challenge.
    Code is available at project homepage \href{https://invinciblewyq.github.io/vstream-page}{https://invinciblewyq.github.io/vstream-page}.
\end{abstract}
\section{Introduction}
\label{sec:intro}
Recently, natural language is becoming a general interface for various tasks and modalities~\cite{alayrac2022flamingo,li2023blip,multimodal3,zhang2024narrative,wang2022vlmixer,wang2024ponder}
Most existing large video-language models face challenges when performing long video question-answering upon user queries~\cite{llamavid,chatunivi,vistallama,moviechat}.
The main reason is that: visual tokens between consecutive frames are heavy and redundant without effective compression, making it
impossible to save all visual features in limited GPU Memory (VRAM), as well as significantly increasing the decoding latency of language model.
Some previous works solve this by feature compression or pruning~\cite{ye2024voco,ye2024atp,wang2024uni,track2}, and achieve real-time performance~\cite{flashvstream,motionlcm,lu2024manicm}.
To better address this issue, we adopt a hierarchical memory mechanism named STAR Memory, which is proposed recently in Flash-VStream~\cite{flashvstream}.
By further fine-tuning the pretrained weight released by Flash-VStream on the training branch of MovieChat-1K,
we successfully leverage the general video compression and understanding ability of STAR Memory, as well as the domain-specific knowledge of MovieChat-1K.
The strong performance of our model demonstrates the effectiveness of the hierarchical memory method, namely the STAR Memory, in long video question-answering tasks.

\section{Method}

\begin{figure*}[t]
  \centering
  \includegraphics[width=1\linewidth]{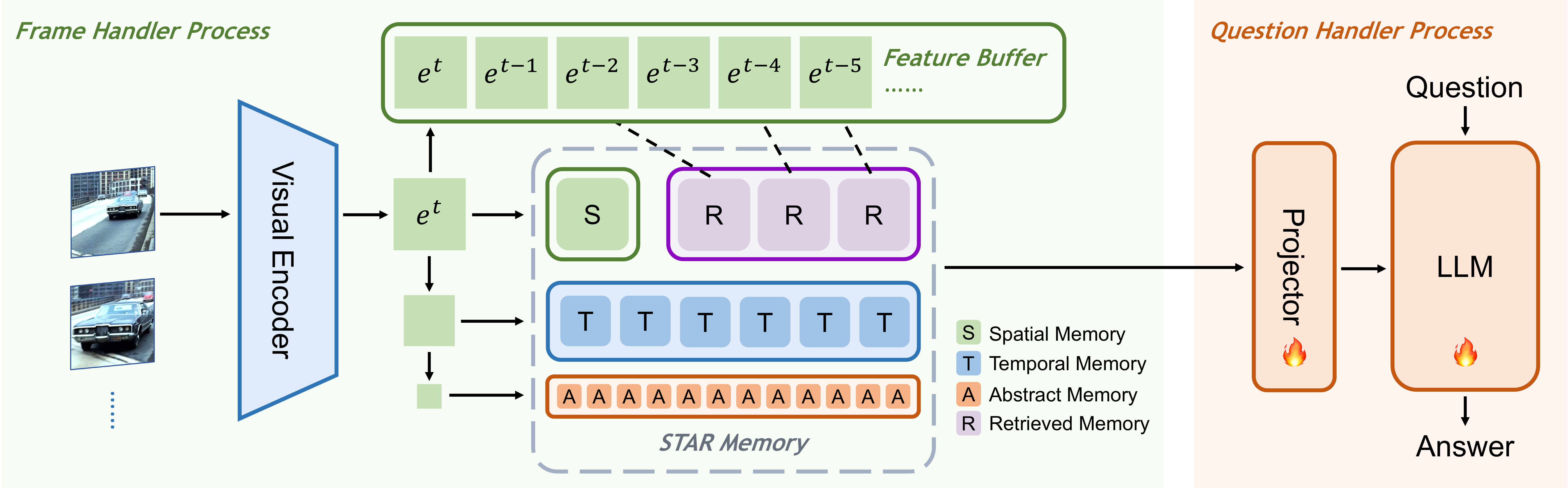}

  \caption{\textbf{The overview of Flash-VStream framework that we adopted for real-time online video stream understanding.}
    Flash-VStream is executed by two processes, namely ``frame handle'' and ``question handler''. The frame handler is responsible for encoding frames and writing to memory, which contains a visual encoder, a STAR memory and a feature buffer. The question handler is responsible for reading from memory and answering questions anytime, which contains a projector and a Large Language Model.
  }
  \label{fig:overall_framework}
\end{figure*}

As shown in ~\Cref{fig:overall_framework}, the Flash-VStream framework that we adopted consists of three main components:
(1) a streaming visual encoder that continuously processes video frames,
(2) a \textbf{S}patial-\textbf{T}emporal-\textbf{A}bstract-\textbf{R}etrieved memory mechanism (\textbf{STAR} memory), including memory writing and reading with the help of a feature buffer.
(3) a LLM decoder capable of providing responses to questions raised by users.

\subsection{Streaming visual encoder}

We use the pre-trained CLIP ViT-L \cite{radford2021learning} as visual encoder. Only patch tokens are used during training and inference.
Specifically, given a frame stream $\{V^t\}_{t=1}^{\infty}$, the encoder maps the $t$-th
frame $V^t\in \mathbb{R}^{H\times W\times 3}$ to feature map ${e^t}\in \mathbb{R}^{P\times P\times D}$, where $P\times P$ is the number of ViT patch tokens and $D$ is the hidden dimension of ViT.

\subsection{\underline{S}patial-\underline{T}emporal-\underline{A}bstract-\underline{R}etrieved memory}
\label{sec:memory}

\begin{figure*}[t]
  \centering
  \includegraphics[width=1.00\linewidth]{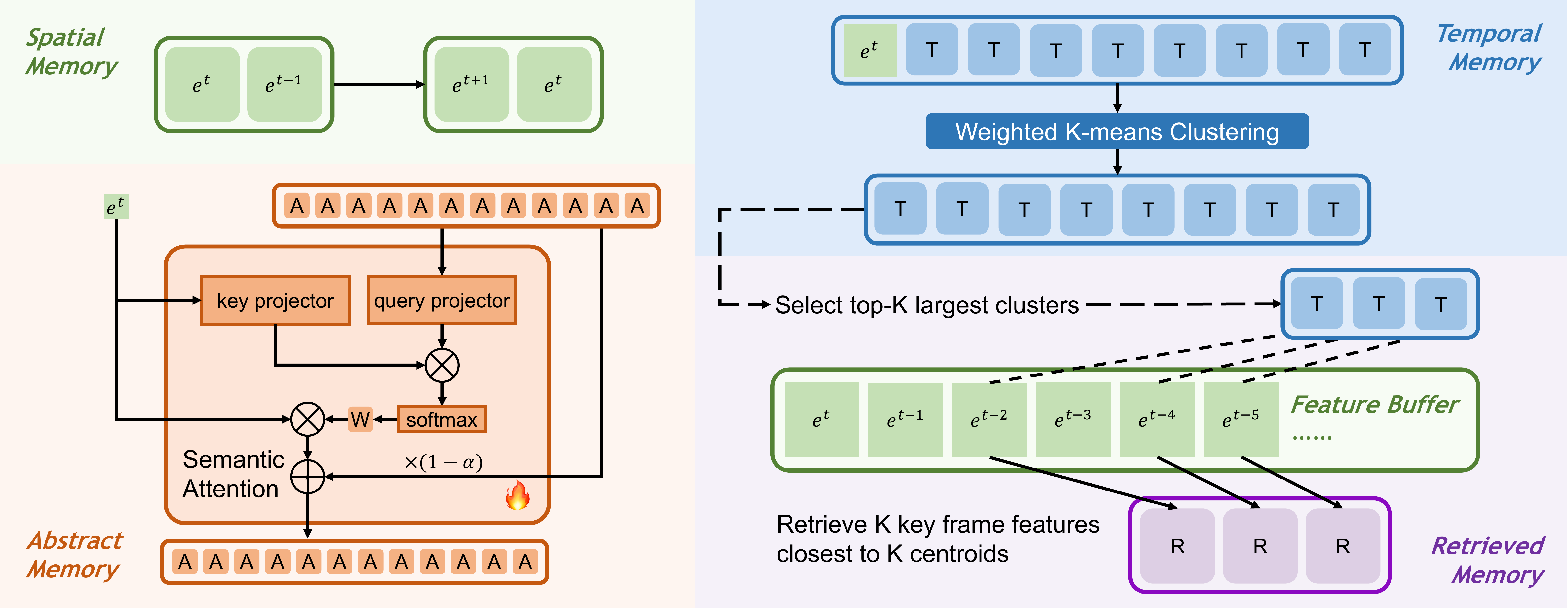} 
  \caption{\textbf{STAR memory writing mechanism.}
    (a) Update spatial memory by a \textit{FIFO} queue.
    (b) Update temporal memory by \textit{Weighted K-means Clustering}.
    (c) Update abstract memory by \textit{Semantic Attention}.
    (d) Update retrieved memory by key frame feature retrival.
    Here feature map $e^T$ has multiple sizes. ``S'', ``T'', ``A'' and ``R'' represent tokens of spatial, temporal, abstract and retrieved memory, respectively.
  }
  \label{fig:memory_mechanism}
\end{figure*}

In order to handle information of different levels of granularity, the STAR memory consists of 4 components:
spatial memory $M_{\text{spa}} \in \mathbb{R}^{N_{\text{spa}} \times P_{\text{spa}}^2 \times D}$,
temporal memory $M_{\text{tem}} \in \mathbb{R}^{N_{\text{tem}}\times P_{\text{tem}}^2 \times D}$,
abstract memory $M_{\text{abs}} \in \mathbb{R}^{N_{\text{abs}}\times P_{\text{abs}}^2 \times D}$ and
retrieved memory $M_{\text{ret}} \in \mathbb{R}^{N_{\text{ret}}\times P_{\text{spa}}^2 \times D}$.
A feature buffer $M_{\text{buff}} \in \mathbb{R}^{N_{\text{buff}}\times P_{\text{spa}}^2 \times D}$ is used to store the feature of latest $N_{\text{buff}}$ frames.
Therefore, the overall memory size is limited to $\text{MAXSIZE}=(N_{\text{spa}}+N_{\text{ret}})\times P_{\text{spa}}^2 + N_{\text{tem}}\times P_{\text{tem}}^2 + N_{\text{abs}}\times P_{\text{abs}}^2$ tokens.

\textbf{Spatial memory.} Spatial memory houses the most recent and detailed spatial information for short-term use, implemented as a \textit{FIFO} (First-In-First-Out) queue, as illustrated in \Cref{fig:memory_mechanism} and \Cref{equ:spa}. This architecture enables continuous updating with the newest frames, facilitating immediate access to fine-grained spatial data.

\textbf{Temporal memory.} Temporal memory integrates dynamic information over time, crucial for long-term retention. When its size surpasses $N_{\text{tem}}$, the $g_{\text{wkmeans}}$ (\textit{Weighted K-means Clustering}) algorithm is applied, as shown in \Cref{equ:tem}. This strategy condenses the memory content into $N_{\text{tem}}$ clusters which can be seen as the representation of key events in videos. Then the centroids of these clusters are used as the new memory for efficiently storing temporal contexts.

\textbf{Abstract memory.} Abstract memory supports high-level semantic concept interpretation through $f_{SA}$, the \textit{Semantic Attention} model. It follows \Cref{equ:abs} to synthesize the insights gained from both spatial and temporal memories into abstracted, actionable knowledge. $f_{SA}$ keeps adjusting $M_{\text{abs}}$, the synopsis of whole video by newest features.
Refer to \Cref{fig:memory_mechanism} for details.

\textbf{Retrieved memory.} Retrieved memory focuses on recalling precise spatial details by identifying and retrieving the most substantial frame features. As shown in \Cref{fig:memory_mechanism}, it first selects the  top-K (where K equals $N_{\text{ret}}$) largest clusters from the $N_{\text{tem}}$ clusters obtained in temporal memory $M_{\text{tem}}$. Then the nearest frame features in feature buffer to centroids of these K clusters are retrieved to supplement the temporal memory with more detailed spatial information. This process is illustrated in \Cref{equ:ret}.

In brief, a new feature $e^t$ is written to STAR memory as follows:
\begin{align}
  M_{\text{buff}}^t & = \text{concat}\big(g_{\text{pooling}}(e^t, P_{\text{spa}}) , M_{\text{buff}}^{t-1}\big) [0:N_{\text{buff}},:,:] \label{equ:buff} \\
  M_{\text{spa}}^t  & = M_{\text{buff}}^t[0:N_{\text{spa}},:,:] \label{equ:spa}                                                                         \\
  M_{\text{tem}}^t  & = g_{\text{wkmeans}}\Big(
  \text{concat}\big(g_{\text{pooling}}(e^t, P_{\text{tem}}) , M_{\text{tem}}^{t-1}\big), N_{\text{tem}}\Big) \label{equ:tem}                            \\
  M_{\text{abs}}^t  & = f_{SA}\big(M_{\text{abs}}^{t-1}, g_{\text{pooling}}(e^t, P_{\text{abs}}), N_{\text{abs}}\big) \label{equ:abs}                   \\
  M_{\text{ret}}^t  & = g_{\text{retrieve}}(M_{\text{buff}}^t, M_{\text{tem}}^t, N_{\text{ret}}) \label{equ:ret}
\end{align}
\vspace{-10pt}

Here $g_{\text{pooling}}(e,P^\prime)$ applies \textit{Average Pooling} to compress feature map $e$ from $P^{2}$ to $P^{\prime 2}$ size along width and height dimensions. $\texttt{concat}(a,b)$ means concatenating tensors $a$ and $b$ along time axis.

\subsection{Real-time LLM decoder}
The LLM decoder works as part of a real-time question answering server. When triggered by a question $Q^t$ at time $t$, the LLM decoder first calculates the text embedding $I_{\text{text}}^t = f_{\text{embed}}(Q^t)$ and maps the STAR memory $M^t=M_{\text{spa}}^t+M_{\text{tem}}^t+M_{\text{abs}}^t+M_{\text{ret}}^t$ to embedding space with the projector $I_{\text{vision}}^t = f_{\text{proj}}(M^t)$. Then it starts to generate answer $A^t = f_{\text{LLM}}(I_{\text{text}}^t, I_{\text{vision}}^t).\text{decode}()$ in real time.

\subsection{Adopting automatic speech recognition (ASR)}
To further utilize the audio information in videos, we adopt the automatic speech recognition (ASR) model to transcribe the audio stream into text.
The transcribed text is then fed into the LLM decoder as an additional input to improve the performance of video question answering.

\subsection{Implementation details}
In this study, we utilize pre-trained CLIP~ViT-L/14-336px~\cite{radford2021learning} as streaming visual encoder.
Following LLaVA~\cite{liu2023visual}, we choose a 2-layer-MLP as visual projector and pre-trained Vicuna-7B~\cite{vicuna2023} as LLM decoder.
We adopt the open-source ASR model Whisper-large-v3~\cite{whisper} to pre-transcribe the audio stream into text.

Considering the balance between performance and resource consumption, we set $P_{\text{spa}}=8$, $P_{\text{tem}}=4$, $P_{\text{abs}}=1$, $N_{\text{buff}}=300$, $N_{\text{spa}}=1$, $N_{\text{tem}}=N_{\text{abs}}=25$ and $N_{\text{ret}}=3$. The MAXSIZE of STAR memory is set to 681 tokens in order to keep computational efficiency.

We train Flash-VStream for 3 stages: modality alignment, instruction tuning, and domain-specific fine-tuning.
The training data of first 2 stages are keep same with LLaMA-VID~\cite{llamavid}, including LLaVA-filtered-558K \cite{llava} image-caption pairs and LLaMA-VID-filtered-232K \cite{llamavid} video-caption pairs for stage 1, LLaVA-filtered-665K \cite{llava} image QA pairs and Video-ChatGPT-filtered-98K \cite{video-chatgpt} video QA pairs for stage 2.
The training data of stage 3 is the training branch of MovieChat-1K~\cite{moviechat}. Speech data transcribed from the ASR model are only used in the third stage.
Speech data are feed into the LLM decoder as a part of the question text, thus providing additional information for video question answering.

For each stage, the model is trained for 1 epoch on 8 A100 80G GPUs.
During training, the parameters of visual encoder are frozen and the parameters of LLM are frozen only for the first stage.
All training and inference experiments was conducted under BF16 precision to save time and resources.

During inference on the test branch of MovieChat-1K, we use different strategy for global and breakpoint questions.
For global questions, we use the STAR memory to store the whole video information and answer the questions.
For breakpoint questions, we only use a part of the video, which ranges from 15s before the breakpoint to 15s after the breakpoint, to answer the questions.

\section{Experiments}

\begin{table}
  \caption{\textbf{Performance on MovieChat-1K dataset.}
    \textbf{G.} and \textbf{B.} denote global and breakpoint, while \textbf{Acc.} and \textbf{Sco.} denote accuracy and score, respectively. \textbf{FVS} denotes Flash-VStream-7b~\cite{flashvstream}.
  }
  \label{tab:VideoQA}
  \centering
  \fontsize{8.3pt}{9pt}\selectfont
  \setlength{\tabcolsep}{5.5pt}
  \begin{tabular}{lcccc}
    \toprule
    \multirow{2}{*}{\textbf{Method}} & \multicolumn{4}{c}{\textbf{val set, GPT-3.5 eval}}                                                          \\
    \cline{2-5}
    \addlinespace[1.5pt]
    ~                                & \textbf{G. Acc.}                                   & \textbf{G. Sco.} & \textbf{B. Acc.} & \textbf{B. Sco.} \\
    \midrule
    FVS                              & 66.7                                               & 3.823            & 53.9             & 3.518            \\
    \midrule
    \textbf{FVS + stage-3}           & \textbf{84.0}                                      & \textbf{4.624}   & \textbf{73.5}    & \textbf{4.078}   \\
    \bottomrule 
    \\
  \end{tabular}
  
  \begin{tabular}{lcccc}
    \toprule
    \multirow{2}{*}{\textbf{Method}} & \multicolumn{4}{c}{\textbf{test set, Gemini-Pro eval}}                                                          \\
    \cline{2-5}
    \addlinespace[1.5pt]
    ~                                & \textbf{G. Acc.}                                       & \textbf{G. Sco.} & \textbf{B. Acc.} & \textbf{B. Sco.} \\
    \midrule
    MovieChat \cite{moviechat}       & 55.1                                                   & 2.78             & 38.5             & 1.87             \\
    \midrule
    FVS + stage-3                    & 84.5                                                   & 4.12             & 52.4             & 2.89             \\
    \midrule
    \textbf{FVS + stage-3 \& speech} & \textbf{96.0}                                          & \textbf{4.60}    & \textbf{59.6}    & \textbf{2.99}    \\
    \bottomrule
  \end{tabular}

\end{table}

To better showcase how different designs affect the performance of our method,
we split out 20\% of the MovieChat-1K training set as the validation set and evaluate our model on it with GPT-3.5 as the evaluator.
As shown in the upper part of \Cref{tab:VideoQA}, it is necessary to fine-tune the pretrained weight released by Flash-VStream on the training branch of MovieChat-1K,
as the performance of the model is significantly improved after stage-3 fine-tuning.
The lower part of \Cref{tab:VideoQA} shows the performance of our model on the test set of MovieChat-1K with Gemini-Pro as the evaluator, evaluated by the organizers.
The performance of our model is significantly improved after adding the speech data as an additional input to the LLM decoder.
It also surpass the training-free baseline method MovieChat~\cite{moviechat} by a large margin, demonstrating the effectiveness of our method
as well as the importance of audio information when processing MovieChat-1K dataset.

\section{Conclusion}

In conclusion, we adopted Flash-VStream~\cite{flashvstream}, a recently proposed long video-QA model.
By incorporating a hierarchical memory, the model can effectively compress visual tokens throughout the whole video.
We further fine-tuned the pretrained weight released by Flash-VStream on the training branch of MovieChat-1K.
The transcribed speech data is also leveraged as an additional input to the LLM decoder to further improve its performance on the MovieChat-1K dataset,
achieving 1st place in the challenge.
We hope our method could inspire further research and advancements in the field of long video stream understanding.

{
  \small
  \bibliographystyle{ieeenat_fullname}
  \bibliography{main}
}

\end{document}